\documentclass[accepted]{uai2022} 

\usepackage[american]{babel}

\usepackage{natbib} 
\usepackage{mathtools} 
\usepackage{booktabs} 
\usepackage{tikz} 

\title{Reframed GES with a Neural Conditional Dependence Measure}

\author[1]{\href{mailto:<xinwei.shen@connect.ust.hk>?Subject=Your UAI 2022 paper}{Xinwei~Shen}{}}
\author[2]{Shengyu~Zhu}
\author[3]{Jiji~Zhang}
\author[2]{Shoubo~Hu}
\author[2]{Zhitang~Chen}
\affil[1]{%
    Hong Kong University of Science and Technology
}
\affil[2]{%
    Huawei Noah's Ark Lab
}
\affil[3]{%
    Hong Kong Baptist University
  }
 
   \usepackage{amsmath, amssymb, amsthm}
\usepackage{mathrsfs, bm}
\usepackage[title]{appendix}
\usepackage{todonotes}
\usepackage{xcolor}
\usepackage{enumitem,booktabs,subfigure}
\usepackage{comment}
\usepackage{multirow}
\usepackage[utf8]{inputenc} 
\usepackage[T1]{fontenc}
\usepackage{algorithm}
\usepackage{algorithmic}

\definecolor{columbiablue}{rgb}{0.61, 0.87, 1.0}

\newcommand{\revise}[1]{{\textcolor{black}{#1}}}

\def\bbE{\mathbb{E}}
\def\bbR{\mathbb{R}}
\def\bbP{\mathbb{P}}
\def\cA{\mathcal{A}}
\def\cB{\mathcal{B}}

\def\cE{\mathcal{E}}

\def\cG{\mathcal{G}}
\def\cP{\mathcal{P}}

\def\cN{\mathcal{N}}
\def\cU{\mathcal{U}}
\def\cT{\mathcal{T}}
\def\cH{\mathcal{H}}

\def\bD{\mathbf{D}}
\def\bE{\mathbf{E}}
\def\bV{\mathbf{V}}
\def\bT{\mathbf{T}}
\def\bH{\mathbf{H}}
\def\bZ{\mathbf{Z}}

\def\tf{\tilde{f}}
\def\tg{\tilde{g}}

\newcommand{\indpt}{\perp\!\!\!\perp}
\DeclareMathOperator*{\argmin}{argmin}
\def\pa{\mathbf{Pa}}
\def\nd{\mathbf{Nd}}

\def\pto{\overset{p}{\to}}

\theoremstyle{plain}
\newtheorem{theorem}{Theorem}
\newtheorem{proposition}[theorem]{Proposition}
\newtheorem{lemma}[theorem]{Lemma}

\newtheorem{definition}{Definition}

\theoremstyle{remark}

\newtheorem*{remark}{Remark}

\begin{document}
\maketitle

\begin{abstract}
In a nonparametric setting, the causal structure is often identifiable only up to Markov equivalence, and for the purpose of causal inference, it is useful to learn a graphical representation of the Markov equivalence class (MEC).  In this paper, we revisit the Greedy Equivalence Search (GES) algorithm, which is widely cited as a score-based algorithm for learning the MEC of the underlying causal structure. We observe that in order to make the GES algorithm consistent in a nonparametric setting, it is not necessary to design a scoring metric that evaluates graphs. Instead, it suffices to plug in a consistent estimator of a measure of conditional dependence to guide the search. We therefore present a reframing of the GES algorithm, which is more flexible than the standard score-based version and readily lends itself to the nonparametric setting with a general measure of conditional dependence. In addition, we propose a neural conditional dependence (NCD) measure, which utilizes the expressive power of deep neural networks to characterize conditional independence in a nonparametric manner. We establish the optimality of the reframed GES algorithm under standard assumptions and the consistency of using our NCD estimator to decide conditional independence. Together these results justify the proposed approach. Experimental results demonstrate the effectiveness of our method in causal discovery, as well as the advantages of using our NCD measure over kernel-based measures. 
\end{abstract}

\section{Introduction}

 Causal structure learning is a fundamental problem in various disciplines of science, and flexible solutions to this problem have potentially wide applications
\citep{pearl2009causality, koller2009probabilistic, peters2017elements}, e.g., inferring causal relationships among phenotypes \citep{neto2010causal,zhang2015estimation}, and finding causes in earth system sciences \citep{runge2019inferring} and telecommunication networks \citep{ng2019masked}. In many scenarios, it is expensive or even impossible to perform interventions or randomized experiments in order to discover the causal relationships. This limitation inspires the need to infer or at least systematically produce plausible hypotheses of causal structures from purely observational data, which is often known as causal discovery. General assumptions relating the data distribution to the unknown causal structure have been leveraged to make causal discovery feasible, including the well-known Markov condition and faithfulness assumption~\citep{spirtes2000causation}.

Suppose the unknown causal structure can be properly represented by a directed acyclic graph (DAG) over the observed variables. The last one and a half decades have seen a host of results on the identifiability of the causal DAG from the observational data distribution, under various parametric or semi-parametric assumptions~\citep{shimizu2006lingam, Hoyer2009anm, zhang2012pnl,buhlmann2014cam, peters2014equal}. However, in a nonparametric setting, from the observational distribution, the causal structure is known to be identifiable only up to Markov equivalence. Despite this limitation, it remains a worthy task to learn a graphical representation of the Markov equivalence class (MEC), known as a completed partial directed acyclic graph (CPDAG), for a CPDAG usually reveals some valuable causal information and can be used to guide experimental studies. 

Existing methods for causal discovery targeting the CPDAG are roughly categorized into constraint-based and score-based methods. 
The former uses statistical tests to find conditional (in)dependence relationships in the data and use them as constraints to recover the CPDAG that satisfies them. The PC algorithm~\citep{spirtes2000causation} is a well-known exemplar of this approach. 
%
The latter formulates the task as an optimization problem by assigning a score to each candidate graph and searching for the one with the optimal score. Regarding the search and optimization strategy, many algorithms solve a combinatorial optimization problem by performing a greedy search; on the other hand, starting from \citet{zheng2018dags}, much recent work tackles the problem through a continuous optimization~\citep{yu2019dag,lachapelle2019gradient}. While continuous optimization has advantages in scalability, global convergence is hard to guarantee by using gradient-based algorithms without implausible assumptions such as strong convexity, especially when the model involves neural networks. In contrast, some search algorithms can be shown to achieve global optimality in the  large sample limit even with a relatively sparse search space. One of the best-known procedures of this kind is Greedy Equivalence Search (GES)~\citep{chickering2002optimal}. 

\revise{The standard score-based GES algorithm requires a scoring criterion to evaluate each candidate graph. Classical examples include the BIC~\citep{schwarz1978estimating} and the BDeu scores~\citep{geiger1994learning}. However, most score functions are born out of restrictive parametric assumptions on the data distribution which rarely hold for real-world data. When the parametric model is misspecified, which is very common in real data, the optimality of the standard GES with such a score is not guaranteed to reflect the ground truth even in the large sample limit.}

In this paper, we explore a simple strategy to produce a nonparametric GES. We observe that in order to make GES consistent in a nonparametric setting, it is not necessary to design a scoring metric that evaluates graphs as a whole. Instead, it suffices to define a certain criterion to guide the search at each step of the procedure. The approach we consider in this work is to plug in a consistent estimator of a measure of conditional dependence to provide such guidance. The result is a reframed GES algorithm that is more flexible than the standard score-based version and readily lends itself to the nonparametric setting with a general measure of conditional dependence. This avoids potential model misspecification that commonly occurs in score-based methods. On the other hand, although the reframed GES becomes essentially constraint-based, it retains desirable features of the search strategy of GES and performs significantly better in our experiments than paradigmatic constraint-based methods such as PC.

In addition, we propose a measure of conditional dependence based on a characterization of conditional independence from \citet{daudin1980partial} and a novel neural conditional dependence (NCD) estimator which utilizes the expressive power of deep neural networks. 
Many existing nonparametric measures of conditional dependence are based on kernel methods and leverage characterizations in a Reproducing Kernel Hilbert Space (RKHS), e.g., \citet{gretton2005measuring}. However, kernel methods suffer from high computational complexity, preventing them from efficient applications in large scale problems. In contrast, our neural network based approach can benefit from a large sample size without a severe compromise in computational time.

We highlight our main contributions as follows:
\begin{itemize}
	\item We present a reframing of the GES algorithm that can flexibly incorporate a consistent estimator of a general conditional dependence measure.
	\item We propose a neural conditional dependence (NCD) measure, which utilizes the expressive power of deep neural networks.
	\item We provide theoretical guarantees on the correctness of the reframed GES algorithm and the consistency of the NCD estimator to measure conditional dependence under mild conditions, and demonstrate the empirical advantages of the resulting method in causal discovery. 
\end{itemize}

\section{Background and Related Work}
\subsection{Preliminaries and Notations}
Let $\cG=(\bV,\bE)$ be a directed acyclic graph (DAG) consisting of nodes $\bV=(X_1,\dots,X_d)$, each of which is a (possibly multi-dimensional) random variable, and directed edges $\bE$ that connect pairs of nodes. Let $\pa^\cG_i$ be the set of parents of node $X_i$. We denote the joint distribution of $\bV$ by $P_\bV$. A basic problem of causal discovery aims at inferring the unknown causal DAG $\cG$ from an independent and identically distributed (i.i.d.) sample from $P_\bV$. In general, we need assumptions relating the DAG $\cG$ and the distribution $P_\bV$ to make the task possible. A principle adopted by all causal discovery methods is the {\it causal Markov condition}: $P_\bV$ is Markovian with respect to DAG $\cG$, in the sense that every conditional independence statement entailed by $\cG$ according to the standard Markov property of DAGs is true of $P_\bV$. We also assume the commonly adopted {\it faithfulness assumption}: $P_\bV$ is faithful with respect to DAG $\cG$, in the sense that every conditional independence statement true of $P_\bV$ is entailed by $\cG$.
If two DAGs $\cG_1$ and $\cG_2$ entail the same set of conditional independence statements, they are said to be {\it Markov equivalent}. The set of all DAGs that are Markov equivalent to a DAG $\cG$ is called the {\it Markov equivalence class} (MEC) of $\cG$, which can be represented by a completed partially directed acyclic graph (CPDAG). 
For random variables $X,Y$ and $Z$, we write $X\indpt Y\mid Z$ to mean that $X$ and $Y$ are conditionally independent given $Z$. 

\subsection{Related Work on Causal Discovery}


The point that GES can be recast in the spirit of a constraint-based method has been noted in the literature, most recently by \citet{chickering2020statistically} and most explicitly by \citet{nandy2018}. To our knowledge, however, the idea of running GES without a global scoring metric has not been sufficiently explored. In Nandy et al.'s insightful discussion, for example, they emphatically show how a consistent global score can be constructed from local conditional dependence scores in multivariate Gaussian and nonparanormal settings, and stop short of considering the option of dispensing with global scoring altogether. As we aim to demonstrate in this paper, a reframed GES without global scoring is especially flexible and useful in a nonparametric context.  

In our experiments, we use a number of state-of-the-art causal discovery algorithms for comparison, in addition to the aforementioned PC and standard GES. 
One of them is the CAM algorithm~\citep{buhlmann2014cam}, which decouples the search for the causal ordering from the selection of parents for each variable, by leveraging an additive modeling assumption. 
\citet{huang2018generalized} propose a generalized score function (GSF) and apply it in the GES algorithm. Specifically, they transform the statistical decision about conditional independence to a model selection problem for a regression task in an RKHS, define a score based on the penalized log-likelihood for the kernel regression, and then use the score to guide local moves in GES. This work is closely related to ours in that both works are motivated by the goal to develop a nonparametric score to guide the local moves of GES. However, our approach differs from GSF in at least two notable aspects. First, by highlighting the sufficiency of designing a local score that enables consistent statistical decisions about conditional independence, we propose a simpler and more flexible way to dispense with parametric assumptions in GES. 
Second, the specific score we propose is based on neural networks rather than kernels and hence enjoys better computational efficiency when scaling to large sample size. 
Another earlier work sharing a similar spirit is the kernel generalized variance (KGV)~\citep{bach2002learning}, which is also compared in our experiments. 

Other methods follow  \citet{zheng2018dags}, who reformulate the original combinatorial problem into a continuous optimization problem, named NOTEARS, which is solved using the augmented Lagrangian algorithm. Several follow-up works extend NOTEARS to nonlinear causal models, including DAG-GNN~\citep{yu2019dag}, GraN-DAG~\citep{lachapelle2019gradient}, and \citet{ng2019masked}, all of which utilize neural networks to model the nonlinear causal relations. 
In addition, \citet{Zhu2020Causal,ijcai2021-491} adopt policy gradient to search for a DAG with the optimal score.

Since the main purpose of this work is to make GES more applicable in nonparametric settings, in our comparisons we focus mainly on methods that are designed to handle data for continuous variables generated from fairly complex, nonlinear models, and leave out some important methods designed to learn Bayesian networks for discrete variables, such as \cite{bartlett2017integer}.

\subsection{Related Work on Conditional Independence}
Conditional independence plays an important role in many statistics and machine learning problems, ranging from graphical models~\citep{koller2009probabilistic} to invariance learning~\citep{arjovsky2019irm}. 
A number of studies were devoted to characterizing conditional independence or developing conditional independence tests. 
\citet{gretton2005measuring} introduce the Hilbert-Schmidt independence criterion (HSIC), which is extended by \citet{fukumizu2007kernel} to cover conditional independence and used for a conditional independence test. Recently, \citet{azadkia2019simple} propose a surprisingly simple nonparametric measure of conditional dependence based on ranking statistics, which we refer to as the Rank Conditional Dependence (RCD) measure and will use later to illustrate the flexibility of our approach. 
Other works focus on constructing tests of conditional independence by proposing various test statistics, including the kernel conditional independence test~\citep{Zhang2011KernelbasedCI} and the test based on a generalized covariance measure~\citep{shah2020}, among others. 

\section{Reframing the GES Algorithm}
\subsection{Standard GES}
The GES algorithm~\citep{chickering2002optimal} searches over the space of MECs of DAGs, which are represented by CPDAGs. The connectivity in the search space is given by the independence-map (IMAP) relation:  graph $\cG$ is an IMAP of graph $\cG'$ if every conditional independence entailed by $\cG$ is entailed by $\cG'$. The standard GES uses a scoring function that assigns a score to every DAG given data, and uses the score of a representative DAG in a MEC as that for the MEC. The search strategy consists of two phases, a phase of forward equivalence search (FES) followed by a phase of backward equivalence search (BES). In FES, the procedure starts with the empty CPDAG (the one with no edges), and moves at each step to a best-scoring CPDAG with one more adjacency (that is an IMAP of the previous CPDAG), until the score cannot be improved by adding more adjacencies. In BES, the procedure starts with the output from FES, and moves at each step to a best-scoring CPDAG with one fewer adjacency (of which the previous CPDAG is an IMAP), until the score cannot be improved by deleting more adjacencies.

We enter some details of FES to highlight the observation that motivates the subsequent reframing. The case of BES is analogous. In FES, each step considers possible insert-one-edge operations on the current CPDAG. Following \citet{chickering2002optimal}, for non-adjacent nodes $X_i$ and $X_j$ in a CPDAG $\cP$, and for any subset $\bT$ of the neighbors of $X_j$ (i.e., nodes that are connected to $X_j$ by undirected edges) that are not adjacent to $X_i$, the $Insert(X_i,X_j,\bT)$ operator modifies $\cP$ to obtain $\cP'$ by inserting the directed edge $X_i\to X_j$, and for each $T\in\bT$, directing the previously undirected edge between $T$ and $X_j$ as $T\to X_j$. 
If the validity condition in \citet[Theorem~15]{chickering2002optimal} is met, then there is a representative DAG $\cG$ in the MEC represented by $\cP$ and a representative DAG $\cG'$ in the MEC represented by $\cP'$, such that $\cG'$ is the result of inserting $X_i\to X_j$ in $\cG$ (which implies that $\cP'$ is an IMAP of $\cP$).

In Chickering's (\citeyear{chickering2002optimal}) proof of the asymptotic correctness of GES under the causal Markov and faithfulness assumptions, the crucial condition is that the ``local'' decision between $\cG$ and $\cG'$ mentioned above asymptotically tracks whether a certain conditional independence relation holds. We make this notion precise in the following definition, in which $\bD$ denotes an i.i.d. sample with size $n$ from the joint distribution $P_\bV$ of $\bV$.
\begin{definition}[Independence-tracking decision criterion]\label{def:local_cons}
	Let $\cG$ and $\cG'$ be two DAGs over $\bV$ that are exactly the same except that $\cG'$ contains an edge $X_i\to X_j$ that does not appear in $\cG$. A decision criterion (based on data $\bD$) to choose between $\cG$ and $\cG'$ (among other options) is independence- tracking if the following two properties hold in the large sample limit:
\begin{enumerate}[label=(\roman*)]
\item If $X_j\indpt X_i\mid \pa^\cG_j$ (according to $P_\bV$), then the decision criterion favors $\cG$ over $\cG'$.
\item Otherwise, the decision criterion favors $\cG'$ over $\cG$.
\end{enumerate}
\end{definition}

In the standard GES algorithm, a scoring function for DAGs is used to make such local decisions. The induced decision criterion is independence-tracking if the scoring function satisfies the so-called local consistency~\citep[Definition~6]{chickering2002optimal}. Indeed, Definition~\ref{def:local_cons} is a straightforward generalization of the notion of local consistency for scoring functions. The generalization serves to highlight a simple but important observation: the crucial condition for the optimality of GES can be implemented by a (locally consistent) score function for DAGs, but does not necessitate such a function.      

\subsection{Reframed GES}
We now describe a simple alternative way to implement an independence-tracking decision criterion for GES, by using any consistent measure of conditional dependence, in the following sense:


\begin{definition}[$\tau$-consistency]\label{def:score_cons}
Consider a set of statistics $\cT = \{T_n(X, Y|\mathbf{Z})\mid X, Y\in \bV,\mathbf{Z} \subseteq \bV\backslash \{X, Y\}\}$ (intended to measure conditional dependence) depending on the sample $\bD$ with size $n$. $\cT$ is said to be $\tau$-consistent with parameter $\tau>0$ if for every $X, Y\in \bV$ and $\mathbf{Z}\subseteq \bV\backslash \{X, Y\}$, 
the following two conditions hold in the large sample limit:
\begin{enumerate}[label=(\roman*)]
\item If $X\indpt Y\mid \mathbf{Z}$ (according to $P_\bV$), then $T_n(X, Y|\mathbf{Z})<\tau$.
\item Otherwise, $T_n(X, Y|\mathbf{Z})>\tau$.
\end{enumerate}
\end{definition}

For our purpose, the following sufficient condition for the $\tau$-consistency in Definition~\ref{def:score_cons} is useful. All proofs are deferred to Appendix~\ref{app:pf}. 

\begin{proposition}\label{prop:suff_cond}
	Suppose for every $X, Y\in \bV$ and $\mathbf{Z}\subseteq \bV\backslash \{X, Y\}$, $T_*(X,Y|\mathbf{Z})\geq0$ is a quantity depending on $P_\bV$ such that
\begin{equation*}
	T_*(X,Y|\mathbf{Z})=0\text{ if and only if }X\indpt Y\mid \mathbf{Z}.
\end{equation*}
Let $\hat{T}_n(X,Y|\mathbf{Z})$ form a set of statistics indexed by $X, Y\in \bV$ and $\mathbf{Z} \subseteq \bV\backslash \{X, Y\}$. 
If $\hat{T}_n(X,Y|\mathbf{Z})\to T_*(X,Y|Z)$ in probability as $n\to\infty$ for every $X, Y\in \bV$ and $\mathbf{Z}\subseteq \bV\backslash \{X, Y\}$, then there exists $\tau>0$ such that $\{\hat{T}_n(X,Y|\mathbf{Z})\}$ is $\tau$-consistent.
\end{proposition}

This proposition provides a way to construct a $\tau$-consistent set of statistics. One first defines a population quantity that takes the boundary value if and only if the conditional independence in question holds. Then one constructs a consistent estimator for this quantity given an i.i.d. sample. The aforementioned measures including HSIC~\citep{fukumizu2007kernel} and RCD~\citep{azadkia2019simple} were both developed along this line. 
In the next section, we will propose a new measure of conditional dependence based on a neural network implementation.

It is worth noting the essential difference between our defined $\tau$-consistent statistics and conditional independence tests. We note that the condition in Proposition~\ref{prop:suff_cond} indicates that when $X\indpt Y\mid Z$, the statistic $\hat{T}_n(X,Y|\mathbf{Z})$ converges to 0 in probability, i.e., $\hat{T}_n=o_p(1)$. By contrast, in a typical conditional independence test, one usually uses a test statistic that, under the null hypothesis of conditional independence, follows an asymptotic null distribution, which is then used to develop a decision rule. This means that when $X\indpt Y\mid Z$, the test statistic is stochastically bounded, i.e., $O_p(1)$, but not necessarily $o_p(1)$. Therefore, it is in general non-trivial to define a $\tau$-consistent statistic from a conditional independence test. 

With a $\tau$-consistent $\hat{T}(X,Y|\mathbf{Z})$, it is straightforward to implement an independence-tracking decision criterion to be used in GES. Specifically, at each step in FES, to each (valid) operator $Insert(X_i, X_j, \bT)$ we assign $\hat{T}(X_i,X_j|\pa^\cG_j)$ as its ``local score'' (where $\cG$ is the DAG representing the current CPDAG induced by the operator), and apply the operator with the highest local score (indicating conditional dependence), unless all remaining valid insert operators yield a local score lower than the threshold $\tau$. Similarly, at each step in BES, to each (valid) operator $Delete(X_i, X_j, \mathbf{H})$, we assign $\hat{T}(X_i,X_j|\pa^\cG_j)$ as its local score (where $\cG$ is the DAG representing the CPDAG the operator would produce), and apply the operator with the lowest local score (indicating conditional independence), unless all remaining valid delete operators yield a score greater than $\tau$. The update step of FES is summarized in Algorithm \ref{FESupdate}, and the dual update step of BES is given in Appendix~\ref{app:bes}. 

\begin{algorithm}[t]
\small
\caption{The update step in the reframed FES}
\label{FESupdate}
\textbf{Input}: the current CPDAG $\cP$, sample $\bD$, a list of valid insert operators $\mathbf{INS}$, statistics $\hat{T}(X,Y|\mathbf{Z})$, threshold $\tau$ \\
\textbf{Output}: the next CPDAG $\cP'$
\begin{algorithmic}
\STATE Set $s=0$ and $I=\texttt{NULL}$.
\FOR{$Insert(X_i, X_j, \bT)\in \mathbf{INS}$}
\STATE Let $\cG$ be the representative DAG of $\cP$ corresponding to $Insert(X_i, X_j, \bT)$.
\STATE Evaluate $Score(X_i, X_j, \bT)=\hat{T}(X_i,X_j|\pa^\cG_j)$. 
\IF {$Score(X_i, X_j, \bT)>s$}
\STATE Let $s=Score(X_i, X_j, \bT)$ and $I=Insert(X_i, X_j, \bT)$.
\ENDIF
\ENDFOR
\IF {$s>\tau$}
\STATE Apply operator $I$ to obtain $\cP'$.
\ELSE
\STATE Keep $\cP'=\cP$ (and terminate FES).
\ENDIF
\STATE \textbf{return} $\cP'$
\end{algorithmic}
\end{algorithm}

We call the GES algorithm with these update steps the \emph{reframed GES}. Unlike the standard GES, this reframed GES does not optimize a global score for MECs. However, by using a suitable local score for choosing operators to apply (or to stop), the local decision criterion remains independence-tracking, and as a result the asymptotic optimality of the reframed GES algorithm is still guaranteed, as stated in the following theorem. 
\begin{theorem}\label{thm:opt_cges}
Under the causal Markov and faithfulness assumptions, the reframed GES procedure using a $\tau$-consistent $\hat{T}(X,Y|\mathbf{Z})$ recovers the MEC of the true graph in the large sample limit. 
\end{theorem}

\section{Neural Conditional Dependence Measure}
In this section, we propose a novel measure of conditional dependence. 
Let $X$, $Y$, and $Z$ be three random variables taking values in $\bbR^{d_X}$, $\bbR^{d_Y}$, and $\bbR^{d_Z}$, respectively, where $d_X$, $d_Y$, and $d_Z$ are the corresponding dimensions. We assume that their joint distribution is absolutely continuous with respect to Lebesgue measure with density $p_*$ defined on $\bbR^{d_X+d_Y+d_Z}$. The conditional independence between $X$ and $Y$ given $Z$ is defined by $p_*(x,y,z)=p_*(x|z)p_*(y|z)p_*(z)$ for all $x,y,z$ with $p_*(z)>0$~\citep{dawid1979conditional}. 

The following lemma from \citet{daudin1980partial} characterizes the conditional independence, which has given rise to several hypothesis testing methods. 
Let $L^2_Z$, $L^2_{XZ}$, and $L^2_{YZ}$ be the spaces of square integrable functions of $Z$, $(X,Z)$, and $(Y,Z)$, respectively, e.g., $L^2_{XZ}=\{f:\bbR^{d_X+d_Z}\to\bbR\mid\bbE[f(X,Z)^2]<\infty\}$. 
\begin{lemma}[\citet{daudin1980partial}]\label{lem:daudin}
	The random variables $X$ and $Y$ are conditionally independent given $Z$ if and only if
	\begin{equation*}
		\bbE[f(X,Z)g(Y,Z)]=0,
	\end{equation*}
	for all $f\in L^2_{XZ}$ and $g\in L^2_{YZ}$ such that $\bbE[f(X,Z)|Z]=0$ and $\bbE[g(X,Z)|Z]=0$.
\end{lemma}

At the population level, given a ground truth density $p_*$, we propose the following measure of conditional dependence between $X$ and $Y$ given $Z$:
\begin{equation}\label{eq:score_pop}\small
	S(X,Y|Z)=\sup_{f,g}\rho^2(f(X,Z)-h^*(Z),g(Y,Z)-l^*(Z))
\end{equation}
where $f\in L^2_{XZ}$ and $g\in L^2_{YZ}$ are test functions, $h^*(Z)=\bbE[f(X,Z)|Z]$, $l^*(Z)=\bbE[g(Y,Z)|Z]$, and $\rho(X_1,X_2)=\mathsf{cov}(X_1,X_2)/\sqrt{\mathsf{var}(X_1)\mathsf{var}(X_2)}$ denotes the Pearson correlation coefficient of two random variables $X_1$ and $X_2$. 
The reason for using the correlation coefficient rather than the covariance is that after normalization by the variances, the characteristic is bounded between $[-1,1]$. This makes the measure well-defined in a bounded range and the computation of its subsequent sample version numerically stable. 

Based on Lemma~\ref{lem:daudin}, we have the following simple theorem which characterizes the property of the measure \eqref{eq:score_pop} and 
establishes an equivalence condition between the measure and conditional independence. 
\begin{theorem}\label{thm:pop_equiv}
	For all $p_*$, we have $S(X,Y|Z)\in[0,1]$ and $S(X,Y|Z)=0$ if and only if $X\indpt Y\mid Z$.
\end{theorem}

Having defined the measure $S(X,Y|Z)$, we now make the computation tractable. 
We use deep neural network classes to parametrize the test functions $f,g$ and the conditional expectations $h,l$ in \eqref{eq:score_pop}. Formally, we write $f_\theta$, $g_\phi$, $h_\omega$, $l_\psi$, where the subscripts denote the parameters of the corresponding neural networks. We then exploit the approximation 
\begin{equation}\label{eq:score_pop_nn}
    \sup_{\theta,\phi}\rho^2(f_\theta(X,Z)-h_{\omega^*}(Z),g_\phi(Y,Z)-l_{\psi^*}(Z)),
\end{equation}
where $h_{\omega^*}(z)=h^*(z)$ and $l_{\psi^*}(z)=l^*(z)$. 
According to the universal approximation theorem of neural networks~\citep{hornik1989multilayer}, equation~\eqref{eq:score_pop_nn} can approximate the true measure \eqref{eq:score_pop} with arbitrary accuracy by choosing the appropriate network architecture. 
Since here we mainly focus on the statistical property of the estimator proposed below, we ignore the small approximation error (i.e., the gap between \eqref{eq:score_pop_nn} and \eqref{eq:score_pop}) in the analysis for simplicity.

Next, we present a consistent estimator of $S(X,Y|Z)$.  
Let $\bD=\{(x_i,y_i,z_i),i=1,\dots,n\}$ be the collection of i.i.d. copies of $(X,Y,Z)\sim p_*$. Our estimator of $S(X,Y|Z)$ is given by
%
\begin{equation}\label{eq:score_sam}\small
	\hat{S}_n(X,Y|Z)=\sup_{\theta,\phi}\hat{\rho}^2\big(f_\theta(X,Z)-h_{\hat\omega}(Z),g_\phi(Y,Z)- l_{\hat\psi}(Z)\big),
\end{equation}
where $\hat{\rho}$ is the sample correlation coefficient based on data $\bD$, and
\begin{equation}\label{eq:reg_est}\small
\begin{split}
	\hat \omega=\argmin_{\omega}\frac{1}{n}\sum_{i=1}^n (f_\theta(x_i,z_i)-h_\omega(z_i))^2,\\
	\hat \psi=\argmin_{\psi}\frac{1}{n}\sum_{i=1}^n (g_\phi(y_i,z_i)-l_\psi(z_i))^2,
\end{split}
\end{equation}
are the estimators of $\omega^*$ and $\psi^*$.

\begin{remark}\label{rmk:reg_est}
	The estimators in \eqref{eq:reg_est} based on regression come from the fact that $\bbE[f(X,Z)|Z]=\argmin_{h}\bbE[f(X,Z)-h(Z)]^2$, which is proved in Appendix~\ref{app:pf}.
\end{remark}


We call the proposed estimator \eqref{eq:score_sam} the \emph{neural conditional dependence (NCD)} estimator and its population version \eqref{eq:score_pop_nn} the NCD measure. Since \eqref{eq:score_sam} solves a bilevel optimization problem involving regression problems \eqref{eq:reg_est}, we adopt an alternating gradient descent scheme to obtain the NCD estimator. The procedure is summarized in Algorithm~\ref{alg}, where we alternately update the test functions $f_\theta,g_\phi$ and nonlinear regressors $h_\omega,l_\psi$ for $T_t$ and $T_r$ steps, respectively.

\begin{algorithm}[t]
\small
\caption{Computing the NCD score}
\label{alg}
\textbf{Input}: sample $\bD$, horizon $T_t,T_r$, initial $\theta,\phi,\omega,\psi$\\
\textbf{Output}: NCD score
\begin{algorithmic}[1] 
\FOR{$t_t=1,2,\dots, T_t$}
\FOR{$t_r=1,2,\dots, T_r$}
\STATE Update $\omega$ by descending $\sum_i\nabla_\omega(f_\theta(x_i,z_i)-h_\omega(z_i))^2$\\
\STATE Update $\psi$ by descending $\sum_i\nabla_\psi(g_\phi(y_i,z_i)-l_\psi(z_i))^2$
\ENDFOR
\STATE Update $\theta,\phi$ by ascending the gradient of \\\ \ \ \ $\hat{\rho}^2(f_\theta(X,Z)-h_\omega(Z),g_\phi(Y,Z)-l_\psi(Z))$
\ENDFOR
\STATE Compute $\hat{s}=\hat{\rho}^2(f(X,Z)-h(Z),g(Y,Z)-l(Z))$ 
\STATE \textbf{return} $\hat{s}$
\end{algorithmic}
\end{algorithm}

To study the asymptotic behavior of the proposed estimator, we assume the following regularity conditions, all of which are mild assumptions commonly adopted in the literature. 
\begin{enumerate}[leftmargin=*,label=\textit{C\arabic*}.]
\item The parameter spaces $\theta\in\Theta$, $\phi\in\Phi$, $\omega\in\Omega$, and $\psi\in\Psi$ are compact.
\item $f_\theta(x,z)$, $g_\phi(y,z)$, $h_\omega(z)$, and $l_\psi(z)$ are continuous with respect to the corresponding parameters and data $x,y,z$.
\item $f_\theta(x,z)$, $g_\phi(y,z)$, $h_\omega(z)$, and $l_\psi(z)$ are dominated square integrable, i.e., there exists a dominating function $F(x,z)$ such that $|f_\theta(x,z)|\leq F(x,z)$ for all $\theta$ and $\bbE[F(X,Z)]^2<\infty$. 
\item For all $\theta,\phi$, there exist unique $\omega^*(\theta)\in\Omega$ and $\psi^*(\phi)\in\Psi$ such that $h_{\omega^*}(z)=h^*(z)$ and $l_{\psi^*}(z)=l^*(z)$ almost surely, respectively. 
\end{enumerate}
The following theorem establishes the consistency of $\hat{S}_n(X,Y|Z)$ as an estimator of the population measure $S(X,Y|Z)$.
\begin{theorem}\label{thm:nci_cons}
	Under the regularity conditions \textit{C1-C4}, as $n\to\infty$, we have $\hat{S}_n(X,Y|Z)\to S(X,Y|Z)$ in probability.
\end{theorem}


Finally, we apply the NCD measure to causal discovery through the reframing of GES in the previous section. 
We plug the estimator $\hat{S}_n$ into the reframed GES procedure as the ``local score''. 
Based on Theorems~\ref{thm:pop_equiv} and~\ref{thm:nci_cons}, by applying the sufficient condition in Proposition~\ref{prop:suff_cond}, we know that $\hat{S}_n(X,Y|\mathbf{Z})$ satisfies the $\tau$-consistency in Definition~\ref{def:score_cons} with some $\tau>0$. Then Theorem~\ref{thm:opt_cges} implies the asymptotic correctness of our method to recover the true MEC. 


In addition, to demonstrate the flexibility of our reframed GES algorithm in incorporating various conditional dependence measures, we will also test a version of our procedure using the RCD measure recently proposed by \citet{azadkia2019simple}, because it is very easy to compute. The RCD estimator can be shown to satisfy the $\tau$-consistency and hence suits our framework well. For completeness, we provide a description of RCD in Appendix~\ref{app:rcd}. 

\section{Experiments}

\begin{table*}[h]
    \centering
    \begin{tabular}{lllllllll}
    \toprule
        Setting & \multicolumn{2}{c}{GP (1k)} & \multicolumn{2}{c}{GP (5k)} & \multicolumn{2}{c}{MULT (1k)} & \multicolumn{2}{c}{MULT (5k)}  \\
        \cmidrule(r){2-3} \cmidrule(r){4-5} \cmidrule(r){6-7} \cmidrule(r){8-9}
        Methods & SHD & F1 score & SHD & F1 score & SHD & F1 score & SHD & F1 score\\ 
        \midrule
        NCD   & \textbf{5.6$\pm$2.5} & \textbf{0.63$\pm$0.14} & \textbf{4.2$\pm$2.3} & \textbf{0.71$\pm$0.14} & 6.2$\pm$2.9 & 0.59$\pm$0.08 & 5.6$\pm$2.4 & 0.60$\pm$0.08  \\
        RCD   & 9.0$\pm$0.7 & 0.41$\pm$0.07 & 8.4$\pm$1.1 & 0.53$\pm$0.08 & 7.4$\pm$2.1 & 0.51$\pm$0.09 & \textbf{3.2$\pm$1.3} & \textbf{0.67$\pm$0.07} \\
        PC & 8.8$\pm$1.6 & 0.36$\pm$0.15 & 7.2$\pm$2.4 & 0.50$\pm$0.16 & 7.6$\pm$1.7 & 0.44$\pm$0.15 & 4.6$\pm$1.8 &0.57$\pm$0.13 \\
        BIC & 7.0$\pm$2.8 & 0.49$\pm$0.20 & 6.0$\pm$2.5 & 0.59$\pm$0.17 & 4.2$\pm$2.9 & 0.65$\pm$0.09 & 4.4$\pm$3.4 & {0.62$\pm$0.11} \\
        KGV & 8.5$\pm$1.1 & 0.37$\pm$0.08 & 7.5$\pm$0.5 & 0.51$\pm$0.06 & 9.0$\pm$1.9 & 0.35$\pm$0.14 & 7.2$\pm$0.7 & 0.47$\pm$0.07\\
        CAM & 6.0$\pm$3.5 & 0.50$\pm$0.26 & 7.2$\pm$3.7 & 0.52$\pm$0.22 & 10.8$\pm$1.8 & 0.09$\pm$0.07 & 11.2$\pm$2.3 & 0.13$\pm$0.15 \\
        NOTEARS & 11.4$\pm$0.9 & 0.06$\pm$0.08 & 11.6$\pm$0.9 &0.06$\pm$0.08 & 24.8$\pm$3.8 &0.36$\pm$0.07 & 23.6$\pm$4.7 &0.37$\pm$0.07 \\
        DAG-GNN & 11.0$\pm$1.7 &0.00$\pm$0.00 & 11.4$\pm$1.8 &0.03$\pm$0.07 & 16.4$\pm$2.6 &0.37$\pm$0.13 & 13.6$\pm$3.4 &0.40$\pm$0.10\\
        GraN-DAG & 10.6$\pm$1.1 &0.05$\pm$0.06 & 12.2$\pm$1.8 & 0.12$\pm$0.04 & 8.6$\pm$2.6 &0.54$\pm$0.12 & 10.2$\pm$1.9 &0.51$\pm$0.08 \\
        GSF & 6.4$\pm$3.5 & 0.55$\pm$0.19 & \multicolumn{2}{c}{>12h}  & \textbf{3.0$\pm$1.1} & \textbf{0.67$\pm$0.06} & \multicolumn{2}{c}{>12h}  \\
        \bottomrule
    \end{tabular}
    \caption{SHD and F1 score on PNL data sets with 10 nodes, 2 expected degrees, and 1000 and 5000 samples.}\label{tab:pnl2_shd_f1}
\end{table*}

\begin{table*}[h]
    \centering
    \begin{tabular}{lllllllll}
    \toprule
        Setting & \multicolumn{2}{c}{GP (1k)} & \multicolumn{2}{c}{GP (5k)} & \multicolumn{2}{c}{MULT (1k)} & \multicolumn{2}{c}{MULT (5k)}  \\
        \cmidrule(r){2-3} \cmidrule(r){4-5} \cmidrule(r){6-7} \cmidrule(r){8-9}
        Methods & SHD & F1 score & SHD & F1 score & SHD & F1 score & SHD & F1 score\\ 
        \midrule
        NCD   & \textbf{28.4$\pm$3.6} & \textbf{0.55$\pm$0.05} & \textbf{24.6$\pm$3.8} & \textbf{0.58$\pm$0.08} & \textbf{29.2$\pm$4.6} & \textbf{0.52$\pm$0.07} & 29.8$\pm$5.1 & \textbf{0.57$\pm$0.09}  \\
        RCD   & 32.8$\pm$2.2 & 0.39$\pm$0.11 & 32.6$\pm$4.7 & 0.44$\pm$0.10 & 31.4$\pm$4.5 & 0.43$\pm$0.14 & \textbf{27.2$\pm$3.3} & 0.54$\pm$0.06 \\
        PC & 37.6$\pm$1.3 & 0.18$\pm$0.09 & 36.0$\pm$2.7 & 0.26$\pm$0.05 & 36.2$\pm$1.8 &0.23$\pm$0.06 & 34.4$\pm$1.5 & 0.32$\pm$0.08 \\
        BIC & 33.0$\pm$2.1 & 0.45$\pm$0.06 & 30.8$\pm$3.3 & 0.50$\pm$0.09 & 30.8$\pm$5.6 & 0.43$\pm$0.09 & 35.0$\pm$3.9 & 0.39$\pm$0.07 \\
        KGV & 37.8$\pm$0.7 & 0.20$\pm$0.08 & 34.2$\pm$3.9 & 0.33$\pm$0.07 & 37.2$\pm$1.6 & 0.27$\pm$0.02 & 37.4$\pm$2.2 & 0.31$\pm$0.06 \\
        CAM & 33.0$\pm$5.6 & 0.42$\pm$0.13 & 30.6$\pm$3.4 & 0.50$\pm$0.11 & 35.2$\pm$2.8 &0.25$\pm$0.07 & 34.4$\pm$6.5 & 0.31$\pm$0.15 \\
        NOTEARS & 38.8$\pm$1.9 & 0.13$\pm$0.05 & 38.4$\pm$1.8 &0.13$\pm$0.05 & 39.0$\pm$1.6 & 0.33$\pm$0.04 & 39.0$\pm$1.9 & 0.34$\pm$0.07 \\
        DAG-GNN & 39.2$\pm$1.3 &0.03$\pm$0.02 & 39.2$\pm$2.3 &0.05$\pm$0.09 & 37.8$\pm$2.4 &0.26$\pm$0.10 & 39.6$\pm$1.1 &0.25$\pm$0.12 \\
        GraN-DAG & 34.0$\pm$7.9 &0.18$\pm$0.09 & 35.4$\pm$6.9 &0.30$\pm$0.13 & 37.4$\pm$3.2 &0.20$\pm$0.08 & 37.0$\pm$3.5 &0.27$\pm$0.09 \\
        GSF & 34.0$\pm$3.0 & 0.39$\pm$0.05 & \multicolumn{2}{c}{>12h}  & 31.6$\pm$3.2 & 0.38$\pm$0.09 & \multicolumn{2}{c}{>12h} \\
        \bottomrule
    \end{tabular}
    \caption{SHD and F1 score on PNL data sets with 10 nodes, 8 expected degrees, and 1000 and 5000 samples.}\label{tab:pnl8_shd_f1}
\end{table*}


In this section, we compare our proposed method with various existing state-of-the-art causal discovery approaches on both synthetic and pseudo-real data sets.
Baseline methods include score-based methods using BIC~\citep{chickering2002optimal}, KGV~\citep{bach2002learning}, and GSF~\citep{huang2018generalized}; a constraint-based method, PC algorithm~\citep{spirtes2000causation}; a method based on structural causal model, CAM~\citep{buhlmann2014cam}; as well as the emerging methods in the continuous optimization paradigm including NOTEARS~\citep{zheng2018dags}, DAG-GNN~\citep{yu2019dag}, and GraN-DAG~\citep{lachapelle2019gradient}. 
The details of the experimental settings and hyperparameters (including the choice of $\tau$) of baseline methods and ours are given in Appendix~\ref{app:detail}.\footnote{Our code is available at \url{https://github.com/xwshen51/GES-NCD}.}

The causal discovery performance is evaluated using three metrics: the structural hamming distance (SHD), the structural interventional distance (SID)~\citep{peters2015structural} and the F1 score. 
Since our method and many baseline approaches return a CPDAG representing an MEC, both SHD and SID are evaluated between the learned and ground-truth CPDAGs. 
Then the SHD is the smallest number of edge additions, deletions, and reversals to convert the estimated CPDAG into the true CPDAG. The SID counts the number of pairs $(X_i,X_j)$ such that the interventional distribution $p(x_j|do(X_i=x))$ would be miscalculated if we chose the parent adjustment set from the estimated graph. 
We report the SIDs corresponding to the best and worst DAG in the learned MEC. 
The SHD and SID are computed using functions corresponding to CPDAGs in the Causal Discovery Toolbox \citep{kalainathan2020causal}.
The F1 score is defined as the harmonic mean of the precision and the recall. Computing the F1 score involves summarizing the number of correctly estimated edges. Directed edges in the ground-truth CPDAG are deemed correctly estimated if the learned CPDAG contains exactly the same directed edge and are deemed incorrectly otherwise. Undirected edges in the ground-truth CPDAG are converted to two directed edges in the adjacency matrix. When the learned CPDAG contains exactly the same undirected edge, both converted directed edges are correctly estimated. One directed edge and no edge in the learned CPDAG are deemed as correctly estimating 1 and 0 edge, respectively. 
In general, a lower SHD or SID and a higher F1 score indicate a better estimate.


\subsection{Synthetic Data}

As mentioned in previous sections, when a data set does not satisfy the additive Gaussian noise assumption, many existing methods such as BIC, CAM, NOTEARS, and GraN-DAG suffer from model misspecification and thus may lead to misleading results. In contrast, nonparametric methods like GSF and ours in principle will not be affected. Here we consider the well-known post nonlinear (PNL) causal models~\citep{zhang2012pnl}. A general PNL model expresses each variable $X_i$ as $$X_i=g_{i,2}(g_{i,1}(\pa_i)+N_i),\ i=1,\dots,d,$$ where $\pa_i$ contains the direct causes of $X_i$, $N_i$ is the exogenous noise variable, and $g_{i,1}$ and $g_{i,2}$ are nonlinear transformations. 

To synthesize a data set, we first randomly generate a ground-truth DAG $\cG$ following the Erd\H{o}s-R\'enyi (ER) graph model and then generate data following $\cG$ and two types of PNL models that were also considered in \citet{lachapelle2019gradient}. The first one, called \emph{PNL-GP}, samples $g_{i,1}$ independently from a Gaussian process with bandwidth one, takes $g_{i,2}$ as the sigmoid function, and $N_i\sim Laplace(0,b_i)$ with $b_i\sim \cU[0,1]$. All root variables in PNL-GP are sampled from $\cU[-1,1]$. The second one, named \emph{PNL-MULT}, takes $g_{i,1}(x)=\log(sum(x))$ where $sum(x)$ takes the sum of all components of a vector $x$, $g_{i,2}(\cdot)=\exp(\cdot)$, and $N_i\sim|\cN(0,\sigma^2_i)|$ with $\sigma^2_i\sim\cU[0,1]$. All root variables in PNL-MULT are sampled from $\cU[0,2]$. This model is adapted from \citet{zhang2015estimation}.

Tables \ref{tab:pnl2_shd_f1} and \ref{tab:pnl8_shd_f1} present the results of SHD and F1 score on sparse and dense graphs with 10 nodes respectively, where the error bars represent the standard deviations across 5 data sets per setting. The results of SID are basically consistent, which are deferred to Appendix~\ref{app:add_exp} due to the space limit. Additional results on graphs with 20 nodes are also presented in Appendix~\ref{app:add_exp}. We see that in general, the reframed GES algorithm with our own NCD or the adopted RCD (shown in the first two lines of all tables) performs the best across all settings, except in the sparse PNL-MULT data where GSF is the best. The advantages of our methods on the more challenging dense graphs are more significant than those on sparse ones. 
In most cases, NCD outperforms RCD, though RCD produces excellent results on PNL-MULT with a larger sample. From the perspective of implementation, RCD may be favored over NCD in terms of fewer hyperparameters and less computational cost. 
In addition, our methods exhibit similarly good performances across different ground-truth models, while most other methods tend to perform well on at most one setting, which indicates the robustness of our nonparametric approach against different distributions. 

GSF, as another kernel-based nonparametric score, performs very well on PNL-MULT with a sparse structure, but is less competitive in other settings. 
Note that we only report the results of GSF using 1000 samples, because even for the sparse graph, it takes around 17 hours for a single run with 5000 samples compared to around 19 minutes with 1000 samples. In contrast, our NCD-based method can benefit from a larger sample size while taking similar computational time as with a smaller sample (both within 4 minutes in the sparse case). \revise{In Appendix~\ref{app:add_exp}, we discuss more details regarding the computational time of different methods.} 
KGV leads to inferior performance in all settings. 
The standard GES with the linear-Gaussian BIC score sometimes performs well on PNL-MULT; a possible reason is that when the noise variance $\sigma_i^2\sim\cU[0,1]$ happens to be small, the PNL-MULT model behaves similarly to a linear-Gaussian model, leading to a case with minor misspecification for BIC. 
This may also partly account for the fact that PC performs better on PNL-MULT than on PNL-GP; that is, a PNL-MULT data set can be similar to linear-Gaussian data which would satisfy the model assumption made in the hypothesis testing. 
CAM performs better on PNL-GP than on PNL-MULT and achieves the best SID in one case, as shown in Appendix~\ref{app:add_exp}. 
The continuous optimization methods are inferior on these PNL data sets, which could be explained by their misspecification of the model. 

In addition, we evaluate our methods in a multi-dimensional scenario where each node may have more than one dimension. Note that our NCD estimator can be readily applied to the multi-dimensional setup by adjusting the input dimension of the test functions, while the rank-based RCD measure unfortunately cannot be directly applied here. Some of the baseline methods, including CAM, NOTEARS, and GraN-DAG do not apply to the multi-dimensional case, so we do not compare with them in this setting. 
We use 10 synthetic data sets from \citet{huang2018generalized}, each with 5 nodes and a sample size of 1000. As shown in Table \ref{tab:multdim}, our approach outperforms the baseline methods in all three metrics. PC, KGV, and GSF are the second-best performing methods in terms of SHD, SID, and F1 score, respectively, though they all give an inferior performance in other metrics.

\begin{table}
\centering
\begin{tabular}{@{}llll@{}}
\toprule
{Method} & \multicolumn{1}{c}{SHD} & \multicolumn{1}{c}{SID} & \multicolumn{1}{c}{F1 score}\\\midrule
NCD & \textbf{2.6$\pm$2.9} & \textbf{[2.0$\pm$3.7, 14.2$\pm$2.8]} & \textbf{0.73$\pm$0.15} \\
PC & 3.0$\pm$1.3 & [7.4$\pm$4.3, 14.9$\pm$3.7] & 0.57$\pm$0.11 \\
KGV & 4.1$\pm$1.6 & [3.7$\pm$3.3, 17.8$\pm$2.8] & 0.58$\pm$0.11 \\
DAG-GNN\hspace{-0.1in} &4.3$\pm$2.3 &	[4.1$\pm$3.9,	15.6$\pm$4.1]&	0.59$\pm$0.27 \\
GSF & 4.7$\pm$3.0 & [4.3$\pm$4.2, 15.6$\pm$3.0] & 0.61$\pm$0.17 \\
BIC & 4.7$\pm$0.9 & [4.8$\pm$3.8, 16.2$\pm$2.7] & 0.57$\pm$0.00 \\
\bottomrule
\end{tabular}
\caption{Results on 10 multi-dimensional data sets.}\label{tab:multdim}
\end{table}

\subsection{Pseudo-real Data}

Although the synthetic data sets from PNL models can expose the model misspecification problem in many existing methods, they differ from the additive noise setup only by the nonlinearity $g_{i,2}$, and hence amount to relatively mild cases of misspecification. In this section, we consider a pseudo-real data set sampled from the SynTReN generator~\citep{van2006syntren} where there is no guarantee at all for model specification. We evaluate on the 10 data sets sampled by \citet{lachapelle2019gradient}, each with 20 nodes and a small sample size of 500. 
In addition, we consider a real Bayesian network, CHILD network (with 20 nodes), and randomly generate 3000 samples following the PNL-GP model introduced in the previous section. 

As shown in Table \ref{tab:syntren}, on SynTReN, most baseline methods perform poorly, indicating a potentially severe violation of their model assumptions. Our reframed GES with NCD and with RCD obtain the best SHDs. 
The results on SynTReN suggest the potential advantage of nonparametric causal discovery methods in real applications where model misspecification is common and possibly grave. 
Moreover, our methods also obtain the best performance on the real graph structure CHILD. Note that on this large data set with 3000 samples,  the kernel-based methods GSF and KGV face serious computational challenges in that they take more than 12 hours for a single run. Therefore, we expect our method to exhibit even more advantages than the kernel-based methods in large-scale scenarios. 



\begin{table}
\centering
\begin{tabular}{lll}
\toprule
Method & \multicolumn{1}{c}{SynTReN} & \multicolumn{1}{c}{CHILD} \\\midrule
NCD & \textbf{30.0$\pm$5.8} & \textbf{16.8$\pm$3.3} \\
RCD & \textbf{30.9$\pm$4.8} & \textbf{14.0$\pm$5.8} \\
PC & 37.4$\pm$4.1 & 22.6$\pm$9.4 \\
BIC & 65.8$\pm$10.8 & 32.8$\pm$16.8 \\
NOTEARS & 99.8$\pm$14.4 & 23.6$\pm$1.9 \\
DAG-GNN &38.5$\pm$5.1 &	28.8$\pm$3.2 \\
GraN-DAG & 58.7$\pm$10.0	& {17.2$\pm$2.1} \\
KGV & 39.9$\pm$8.0 & >12h \\
GSF & 52.1$\pm$8.9 & >12h\\
\bottomrule
\end{tabular}
\caption{SHD on pseudo-real data.}\label{tab:syntren}
\end{table}

\section{Conclusion}
In this work, we presented a reframed GES algorithm that works with a measure of conditional dependence rather than a scoring metric for graphs. This way the algorithm is easily applicable in a nonparametric setting with a theoretical guarantee. We also proposed a neural conditional dependence (NCD) measure based on a deep neural network implementation, and established its theoretical properties that make it suitable for the reframed GES. The resulting causal discovery algorithm was shown in our experiments to be superior or competitive in comparison to a number of state-of-the-art methods. It also enjoys a significant advantage over kernel-based nonparametric methods in large-scale settings, since the latter are usually infeasible when the sample size is relatively large. For future work, we plan to explore a continuous optimization formulation of causal discovery based on such nonparametric conditional dependence measures. 

\begin{acknowledgements} 
%
JZ’s research was supported in part by the RGC of Hong Kong under GRF13602720 and a start-up fund from HKBU.~
\end{acknowledgements}
The authors thank the anonymous reviewers for their valuable comments and suggestions.

\bibliography{uai2022-template}

\onecolumn
\newpage
\appendix
\section{Reframed BES}\label{app:bes}

We present the dual update step of the reframed BES in Algorithm \ref{BESupdate}.

{\centering
\begin{minipage}{.9\linewidth}
\begin{algorithm}[H]
\caption{The update step in the reframed BES}
\label{BESupdate}
\textbf{Input}: the current CPDAG $\cP$, sample $\bD$, a list of valid delete operators $\mathbf{DEL}$, statistics $\hat{T}(X,Y|\mathbf{Z})$, threshold $\tau$ \\
\textbf{Output}: the next CPDAG $\cP'$
\begin{algorithmic}[1] 
\STATE Set $s=0$ and $I=\texttt{NULL}$.
\FOR{$Delete(X_i, X_j, \bH)\in \mathbf{DEL}$}
\STATE Let $\cG$ be the DAG induced by the operator $Delete(X_i, X_j, \bH)$ that is a representative of the CPDAG the operator would produce.
\STATE Evaluate $Score(X_i, X_j, \bH)=\hat{T}(X_i,X_j|\pa^\cG_j)$. 
\IF {$Score(X_i, X_j, \bH)<s$}
\STATE Let $s=Score(X_i, X_j, \bH)$ and $I=Delete(X_i, X_j, \bH)$.
\ENDIF
\ENDFOR
\IF {$s<\tau$}
\STATE Apply operator $I$ to obtain $\cP'$.
\ELSE
\STATE Keep $\cP'=\cP$ (and terminate BES).
\ENDIF
\STATE \textbf{return} $\cP'$
\end{algorithmic}
\end{algorithm}
\end{minipage}
\par
}

\section{Proofs}\label{app:pf}

\subsection{Proof of Proposition~\ref{prop:suff_cond}}

Define the following two sets of tuples
\begin{equation*}
	\cA=\{(X,Y,\bZ):X,Y\in\bV,Z\subseteq\bV\text{ such that }X\indpt Y\mid\bZ \};
\end{equation*}
\begin{equation*}
	\cB=\{(X,Y,\bZ):X,Y\in\bV,Z\subseteq\bV\text{ such that }X\not\!\indpt Y\mid\bZ\}.
\end{equation*}
We know from the proposition condition that for every $(X,Y,\bZ)\in\cA$, $T_*(X,Y|\bZ)=0$ and for every $(X,Y,\bZ)\in\cB$, $T_*(X,Y|\bZ)>0$. For the number of nodes is finite, the cardinalities of both sets are finite. Then we know $m_0=\min_{(X,Y,\bZ)\in\cB}T_*(X,Y|\bZ)>0$.  Let $\tau$ be any number in $(0,m_0)$. 

 In case (1), by the consistency of $\hat{T}_n$ to $T_*$, we have $\bbP(\hat{T}_n(X,Y|\bZ)>\tau)\to0$ as $n\to\infty$.

 In case (2), we have as $n\to\infty$, $\hat{T}_n(X,Y|\bZ)\pto T_*(X,Y|\bZ)\leq m_0$, where $\pto$ stands for converging in probability, which means for all $\epsilon>0$, $\bbP(|\hat{T}_n(X,Y|\bZ)-T_*(X,Y|\bZ)|<\epsilon)\to1$. By the arbitrariness of $\epsilon$, let $\epsilon<T_*(X,Y|\bZ)-\tau$. Then we have $\{|\hat{T}_n-T_*|<\epsilon\}\subseteq\{T_*-\epsilon<\hat{T}_n\}\subseteq\{\tau<\hat{T}_n\}$. 
 This implies $\bbP(|\hat{T}_n-T_*|<\epsilon)\leq\bbP(\hat{T}_n>\tau)$. Therefore, we have $\bbP(\hat{T}_n>\tau)\to1$ as $n\to\infty$, 
 which concludes the proof.

\subsection{Proof of Theorem~\ref{thm:opt_cges}}

The proof is essentially the same as the proof for the asymptotic correctness of the standard GES with a locally consistent scoring function \citep{chickering2002optimal}, except that the role played by the local consistency of the scoring function is now played by the $\tau$-consistency of $\hat{T}$. We first show that in the large sample limit, the output of the reframed FES is a CPDAG $\cP$ that satisfies the Markov condition with the true distribution $P_\bV$. Suppose for the sake of contradiction that $P_\bV$ is not Markov to $\cP$, which means that $P_\bV$ is not Markov to any DAG $\cG$ in (the MEC represented by) $\cP$.  It follows that there exists a pair of distinct variables $X_i, X_j$ such that they are not adjacent in $\cG$ and $X_i$ is a non-descendant of $X_j$ in $\cG$, but $X_i$ and $X_j$ are not independent given $\pa^\cG_j$ according to $P_\bV$. However, since $\hat{T}$ is $\tau$-consistent, in the large sample limit $\hat{T}(X_i,X_j|\pa^\cG_j)>\tau$, which means that the reframed FES would not have stopped with $\cP$ but would have moved to another CPDAG with an added adjacency between $X_i$ and $X_j$. Contradiction.

Next we show that if the reframed BES starts with a CPDAG that is Markov to $P_\bV$, then in the large sample limit it will output the CPDAG that is both Markov and faithful to $P_\bV$, which represents the true MEC by the causal Markov and faithfulness assumptions. Suppose for the sake of contradiction that the reframed BES ends with a CPDAG $\cP$ that is not faithful to $P_\bV$. Note that $\cP$ would still be Markov to $P_\bV$. If not, since the reframed BES starts with a CPDAG that is Markov to $P_\bV$, there must have been a step where it moved from a CPDAG that is Markov to $P_\bV$ to one that is not. Denote the latter by $\cP'$. It follows that the local score for the operator $Delete(X_i, X_j, \mathbf{H})$ leading to $\cP'$ --- which is equal to $\hat{T}(X_i,X_j|\pa^{\cG'}_j)$, for some $\cG'$ in (the MEC represented by) $\cP'$ --- is smaller than $\tau$ (in the large sample limit) even though $X_i$ and $X_j$ are not independent given $\pa^{\cG'}_j$ according to $P_\bV$. This contradicts the $\tau$-consistency of $\hat{T}$.

Thus the reframed BES ends with a $\cP$ that is Markov but not faithful to $P_\bV$. Let $\mathcal{H}$ denote the true CPDAG, which by assumption is both Markov and faithful to $P_\bV$. Then $\cP $ is an IMAP of $\mathcal{H}$. By Theorem~4 in \citet{chickering2002optimal}, there is a $\cP'$ with one more adjacency than $\cP$ has such that $\cP$ is also an IMAP of $\cP'$. It follows that there is a DAG $\cG'$ representing $\cP'$ and a $\cG$ representing $\cP$ such that $\cG'$ and $\cG$ are the same except for an edge $X_i\rightarrow X_j$ in $\cG'$ but not in $\cG$, and $X_i\indpt X_j \mid \pa^{\cG'}_j$ according to $P_\bV$. Since $\hat{T}$ is $\tau$-consistent, we have $\hat{T}(X_i,X_j|\pa^{\cG'}_j) < \tau$ in the large sample limit. But this means that the reframed BES would not have stopped at $\cP$ but would have continued to some other CPDAG. A contradiction. 

Therefore, the reframed FES followed by the reframed BES will output the true CPDAG $\cH$ 
in the large sample limit.

\subsection{Proof of Theorem~\ref{thm:pop_equiv}}

It is obvious that a correlation coefficient always lies in $[-1,1]$, so $S(X,Y|Z)\in[0,1]$. For the second half of the theorem, note that 
\begin{equation*}
	\rho(f(X,Z)-h^*(Z),g(Y,Z)-l^*(Z))=\frac{\bbE[(f(X,Z)-h^*(Z))(g(Y,Z)-l^*(Z))]}{\sqrt{\bbE[f(X,Z)-h^*(Z)]^2\bbE[g(Y,Z)-l^*(Z)]^2}}.
\end{equation*}
We have
\begin{equation*}
	\{f\in L^2_{XZ}:\bbE[f(X,Z)|Z]=0\}=\{\tilde{f}|\tilde{f}(X,Z)=f(X,Z)-\bbE[f(X,Z)|Z],f\in L^2_{XZ}\}:=\cE_{XZ},
\end{equation*}
\begin{equation*}
	\{g\in L^2_{YZ}:\bbE[g(Y,Z)|Z]=0\}=\{\tilde{g}|\tilde{g}(Y,Z)=g(Y,Z)-\bbE[g(Y,Z)|Z],g\in L^2_{YZ}\}:=\cE_{YZ}.
\end{equation*}
We thus have $S(X,Y|Z)=0$ if and only if 
\begin{equation*}
	\bbE[\tf(X,Z)\tg(Y,Z)]=0\quad\forall \tf\in\cE_{XZ},\tg\in \cE_{YZ}.
\end{equation*}
Then by Lemma~\ref{lem:daudin}, we have $S(X,Y|Z)=0$ if and only if $X\indpt Y\mid Z$.

\subsection{Proof of Theorem~\ref{thm:nci_cons}}

Rewrite the NCD estimator as
\begin{equation*}
	\hat{S}_n=\sup_{\theta\in\Theta,\phi\in\Phi} \frac{\hat{\mathbb{E}}^2[(f_\theta(X,Z)-h_{\hat\omega}(Z))\cdot(g_\phi(Y,Z)-l_{\hat\psi}(Z))]}{\hat{\mathbb{E}}[f_\theta(X,Z)-h_{\hat\omega}(Z)]^2\cdot\hat{\mathbb{E}}[g_\phi(Y,Z)-l_{\hat\psi}(Z)]^2},
\end{equation*}
where $\hat\bbE$ denotes the sample mean given the sample $\bD=\{(x_i,y_i,z_i),i=1,\dots,n\}$, e.g., 
\begin{equation*}
	\hat{\mathbb{E}}[f_\theta(X,Z)-h_{\hat\omega}(Z)]^2=\frac{1}{n}\sum_{i=1}^n [f_\theta(x_i,z_i)-h_{\hat\omega}(z_i)]^2.
\end{equation*}

By the continuous mapping theorem, it suffices to show the following three convergence statements uniformly over $\theta\in\Theta$ and $\phi\in\Phi$:
\begin{enumerate}[leftmargin=*,label=(\roman*)]
\item $\sup_{\theta\in\Theta,\phi\in\Phi}\big|\hat{\mathbb{E}}[(f_\theta(X,Z)-h_{\hat\omega}(Z))\cdot(g_\phi(Y,Z)-l_{\hat\psi}(Z))] - \mathbb{E}[(f_\theta(X,Z)-h_{\omega^*}(Z))\cdot(g_\phi(Y,Z)-l_{\psi^*}(Z))]\big|\pto0$;
\item $\sup_{\theta\in\Theta}\big|\hat{\mathbb{E}}[f_\theta(X,Z)-h_{\hat\omega}(Z)]^2 - \mathbb{E}[f_\theta(X,Z)-h_{\omega^*}(Z)]^2\big|\pto0$;
\item $\sup_{\phi\in\Phi}\big|\hat{\mathbb{E}}[g_\phi(Y,Z)-l_{\hat\psi}(Z)]^2 - \mathbb{E}[g_\phi(Y,Z)-l_{\psi^*}(Z)]^2\big|\pto0$.
\end{enumerate}

\begin{proof}[Proof of (ii) and (iii)]
By the triangular inequality, we have
\begin{equation}\label{eq:trian}
\begin{split}
	&\sup_{\theta\in\Theta}\big|\hat{\mathbb{E}}[f_\theta(X,Z)-h_{\hat\omega}(Z)]^2 - \mathbb{E}[f_\theta(X,Z)-h_{\omega^*}(Z)]^2\big| \\
	\leq&\sup_{\theta\in\Theta}\big|\hat{\mathbb{E}}[f_\theta(X,Z)-h_{\hat\omega}(Z)]^2 - \hat\bbE[f_\theta(X,Z)-h_{\omega^*}(Z)]^2\big|+\sup_{\theta\in\Theta}\big|\hat{\mathbb{E}}[f_\theta(X,Z)-h_{\omega^*}(Z)]^2 - \mathbb{E}[f_\theta(X,Z)-h_{\omega^*}(Z)]^2\big|
\end{split}
\end{equation}
where the second term on the right-hand side vanishes in probability as $n\to\infty$ by applying the uniform law of large numbers \cite[Theorem 2]{jennrich1969asymptotic}. 

We then write the first term as follows:
\begin{align}
&\left|\frac{1}{n}\sum_{i=1}^n\left([f_\theta(x_i,z_i)-h_{\hat\omega}(z_i)]^2- [f_\theta(x_i,z_i)-h_{\omega^*}(z_i)]^2\right)\right|\nonumber\\
=&\left|\frac{1}{n}\sum_{i=1}^n\left[h_{\hat\omega}(z_i)-h_{\omega^*}(z_i)\right]\left[h_{\hat\omega}(z_i)+h_{\omega^*}(z_i)-2f_\theta(x_i,z_i)\right]\right|\nonumber\\
\leq & \left|\frac{1}{n}\sum_{i=1}^n2f_\theta(x_i,z_i)\left[h_{\hat\omega}(z_i)-h_{\omega^*}(z_i)\right]\right| + \left|\frac{1}{n}\sum_{i=1}^n\left[h^2_{\hat\omega}(z_i)-h^2_{\omega^*}(z_i)\right]\right|.\label{eq:bound}
\end{align}

We recall the definitions
\begin{align*}
\omega^*(\theta)&=\argmin_{\omega\in\Omega}\bbE[f_\theta(X,Z)-h_\omega(Z)]^2\\
\hat\omega(\theta)&=\argmin_{\omega\in\Omega}\frac{1}{n}\sum_{i=1}^n[f_\theta(x_i,z_i)-h_\omega(z_i)]^2.
\end{align*}
By the uniform law of large numbers, for all $\theta\in\Theta$, we have as $n\to\infty$ that $$\sup_{\omega\in\Omega}\big|\hat\bbE[f_\theta(X,Z)-h_\omega(Z)]^2-\bbE[f_\theta(X,Z)-h_\omega(Z)]^2\big|\pto0.$$ Further by condition \textit{C4}, we have for all $\theta\in\Theta$, as $n\to\infty$, $\hat\omega(\theta)\pto\omega^*(\theta)$. 
Let $K$ be an arbitrary compact subset of $\bbR^{d_Z}$. 
Because of the compactness of $\Theta$ and the Lipschitz continuity of $h_{\hat\omega(\theta)}(z)$ and $h_{\omega^*(\theta)}(z)$ over $(\theta,z)\in\Theta\times K$, we have $$\sup_{\theta\in\Theta,z\in K}|h_{\hat\omega(\theta)}(z)-h_{\omega^*(\theta)}(z)|\pto0$$ as $n\to\infty$, where $\|\cdot\|$ stands for the Euclidean norm. By the continuous mapping theorem, we have as $n\to\infty$, 
\begin{equation}\label{eq:unif_conv_square}
	\sup_{\theta\in\Theta,z\in K}|h^2_{\hat\omega(\theta)}(z)-h^2_{\omega^*(\theta)}(z)|\pto0.
\end{equation}

Next, we show the second term in \eqref{eq:bound} vanishes in probability.
Given an arbitrary $r>0$, let $B_r=\{z\in\bbR^{d_Z}:\|z\|\leq r\}$. Let $B_r^c=\bbR^{d_Z}\setminus B_r$ be its complement. 
We have
\begin{align}
\sup_{\theta\in\Theta}\left|\frac{1}{n}\sum_{i=1}^n\left[h^2_{\hat\omega}(z_i)-h^2_{\omega^*}(z_i)\right]\right| &\leq \sup_{\theta\in\Theta}\frac{1}{n}\sum_{i=1}^n\left|h^2_{\hat\omega}(z_i)-h^2_{\omega^*}(z_i)\right|\nonumber\\
&= \sup_{\theta\in\Theta}\frac{1}{n}\sum_{i=1}^n\left[\left|h^2_{\hat\omega}(z_i)-h^2_{\omega^*}(z_i)\right|\mathbf{1}_{\{z_i\in B_r\}}+\left|h^2_{\hat\omega}(z_i)-h^2_{\omega^*}(z_i)\right|\mathbf{1}_{\{z_i\in B_r^c\}}\right]\nonumber\\
&\leq \sup_{\theta\in\Theta,z\in B_r}|h^2_{\hat\omega}(z)-h^2_{\omega^*}(z)| + \frac{2}{n}\sum_{i=1}^nH^2(z_i)\mathbf{1}_{\{z_i\in B_r^c\}}\label{eq:decomp},
\end{align}
where the second term in the upper bound \eqref{eq:decomp} comes from the dominated integrable condition in \textit{C3} with a dominating function $H(z)$. 
By taking $n\to\infty$, the first term in \eqref{eq:decomp} vanishes in probability by \eqref{eq:unif_conv_square}, and the second term in \eqref{eq:decomp} becomes $\bbE[H^2(Z)\mathbf{1}_{\{Z\in B_r^c\}}]$. By the dominated convergence theorem, further by letting $r\to\infty$, $\bbE[H^2(Z)\mathbf{1}_{\{Z\in B_r^c\}}]\to0$. Thus, we have as $n\to\infty$
\begin{equation}\label{eq:bound_conv2}
	\sup_{\theta\in\Theta}\left|\frac{1}{n}\sum_{i=1}^n\left[h^2_{\hat\omega}(z_i)-h^2_{\omega^*}(z_i)\right]\right|\pto0.
\end{equation}

Last, we show the first term in \eqref{eq:bound} vanishes in probability. Again, we consider an arbitrary radius $r>0$ and a compact ball $B'_r=\{(x,z)\in\bbR^{d_X+d_Z}:\|(x,z)\|\leq r\}$. Note that $f_\theta(x,z)$ is continuous and hence is uniformly bounded for all $\theta\in\theta$ and $(x,z)\in {B'_r}$. Then
\begin{align*}
&\left|\frac{1}{n}\sum_{i=1}^nf_\theta(x_i,z_i)\left[h_{\hat\omega}(z_i)-h_{\omega^*}(z_i)\right]\right| \\
\leq& \frac{1}{n}\sum_{i=1}^n|f_\theta(x_i,z_i)[h_{\hat\omega}(z_i)-h_{\omega^*}(z_i)]|\mathbf{1}_{\{(x_i,z_i)\in B'_r\}} + \frac{1}{n}\sum_{i=1}^n|f_\theta(x_i,z_i)[h_{\hat\omega}(z_i)-h_{\omega^*}(z_i)]|\mathbf{1}_{\{(x,z)\in {B'_r}^c\}} \\
\leq& M\sup_{\theta\in\Theta,z\in K}|h_{\hat\omega(\theta)}(z)-h_{\omega^*(\theta)}(z)| + \frac{2}{n}\sum_{i=1}^nF(x,z)H(z)\mathbf{1}_{\{(x,z)\in {B'_r}^c\}}
\end{align*}
where $|f_\theta(x,z)|\leq M$ for all $\theta\in\Theta$ and $(x,z)\in B_r'$, and $F(x,z)$ and $H(z)$ are dominating functions for $f_\theta(x,z)$ and $h_\omega(z)$ respectively. 
Similar to the arguments above, we have as $n\to\infty$
\begin{equation}\label{eq:bound_conv1}
	\sup_{\theta\in\Theta}\left|\frac{1}{n}\sum_{i=1}^nf_\theta(x_i,z_i)\left[h_{\hat\omega}(z_i)-h_{\omega^*}(z_i)\right]\right|\pto0.
\end{equation}
Then by combining convergence results \eqref{eq:bound_conv2} and \eqref{eq:bound_conv1} and recalling the upper bounds \eqref{eq:trian} and \eqref{eq:bound}, we have as $n\to\infty$, (ii) holds. Similarly we can show (iii).
\end{proof}

\begin{proof}[Proof of (i)]
By the triangular inequality, we have
\begin{equation}\label{eq:bound1}
\begin{split}
&\sup_{\theta\in\Theta,\phi\in\Phi}\big|\hat{\mathbb{E}}[(f_\theta(X,Z)-h_{\hat\omega}(Z))\cdot(g_\phi(Y,Z)-l_{\hat\psi}(Z))] - \mathbb{E}[(f_\theta(X,Z)-h_{\omega^*}(Z))\cdot(g_\phi(Y,Z)-l_{\psi^*}(Z))]\big|\\
\leq& \sup_{\theta\in\Theta,\phi\in\Phi}\big|\hat{\mathbb{E}}[(f_\theta(X,Z)-h_{\hat\omega}(Z))\cdot(g_\phi(Y,Z)-l_{\hat\psi}(Z))] - \hat\bbE[(f_\theta(X,Z)-h_{\omega^*}(Z))\cdot(g_\phi(Y,Z)-l_{\psi^*}(Z))]\big|\\
&+\sup_{\theta\in\Theta,\phi\in\Phi}\big|\hat\bbE[(f_\theta(X,Z)-h_{\omega^*}(Z))\cdot(g_\phi(Y,Z)-l_{\psi^*}(Z))]\ - \mathbb{E}[(f_\theta(X,Z)-h_{\omega^*}(Z))\cdot(g_\phi(Y,Z)-l_{\psi^*}(Z))]\big|,
\end{split}
\end{equation}
where the second term on the right-hand side vanishes in probability by the uniform law of large numbers. By some calculations we know that the first term of \eqref{eq:bound1} is upper bounded by
\begin{align*}
&\sup_{\theta\in\Theta}\left|\frac{1}{n}\sum_{i=1}^nf_\theta(x_i,z_i)[h_{\hat\omega}(z_i)-h_{\omega^*}(z_i)]\right|+\sup_{\phi\in\Phi}\left|\frac{1}{n}\sum_{i=1}^ng_\phi(y_i,z_i)[l_{\hat\psi}(z_i)-l_{\psi^*}(z_i)]\right|\\
+&\sup_{\theta\in\Theta,\phi\in\Phi}\left|\frac{1}{n}\sum_{i=1}^n\left[h_{\hat\omega}(z_i)l_{\hat\psi}(z_i)-h_{\omega^*}(z_i)l_{\psi^*}(z_i)\right]\right|,
\end{align*}
where all three terms converge to 0 in probability as $n\to\infty$. Therefore, the left-hand side of \eqref{eq:bound1} vanishes in probability, leading to (i).
\end{proof}

\subsection{Proof of the statement in Remark~\ref{rmk:reg_est}}
We recall that $h^*(Z)=\bbE[f(X,Z)|Z]$. The goal is to show that $h^*=\argmin_{h\in L^2_{Z}}\bbE[f(X,Z)-h(Z)]^2$ almost surely.

For all $h\in L^2_{Z}$, we have 
\begin{equation}\label{eq:pf_rmk2}
\begin{split}
	\bbE[f(X,Z)-h(Z)]^2&=\bbE[(f(X,Z)-h^*(Z))+(h^*(Z)-h(Z))]^2\\
	&=\bbE[f(X,Z)-h^*(Z))]^2+\bbE[h^*(Z)-h(Z)]^2+\bbE[(f(X,Z)-h^*(Z))(h^*(Z)-h(Z))].
\end{split}
\end{equation}
Note that the cross term in the second line of \eqref{eq:pf_rmk2} can be simplified using the law of total expectation as follows 
\begin{align*}
	\bbE[(f(X,Z)-h^*(Z))(h^*(Z)-h(Z))]&=\bbE[\bbE[(f(X,Z)-h^*(Z))(h^*(Z)-h(Z))]|Z]\\
	&=\bbE[(h^*(Z)-h(Z))\bbE[f(X,Z)-h^*(Z)|Z]]\\
	&=\bbE[(h^*(Z)-h(Z))(\bbE[f(X,Z)|Z]-h^*(Z))]\\
	&=0.
\end{align*}
Then \eqref{eq:pf_rmk2} becomes
\begin{equation*}
	\bbE[f(X,Z)-h(Z)]^2=\bbE[f(X,Z)-h^*(Z))]^2+\bbE[h^*(Z)-h(Z)]^2\geq\bbE[f(X,Z)-h^*(Z))]^2,
\end{equation*}
where the equality holds if and only if $h(Z)=h^*(Z)$ almost surely.

\section{Rank Conditional Dependence Measure}\label{app:rcd}

In this section, we briefly introduce the RCI and one may refer to \citet{azadkia2019simple} for details. 
Consider a random variable $Y$ and two random vectors $X$ and $Z$, following the joint distribution $p_*$. Let $\mu$ be the law of $Y$. The following quantity measures the degree of conditional dependence of $Y$ and $Z$ given $X$:
\begin{equation*}
	T(X,Y|Z)=\frac{\int\bbE(\mathrm{Var}(\bbP(Y\geq t|X,Z)|Z))d\mu(t)}{\int\bbE(\mathrm{Var}(1_{\{Y\geq t\}}|Z))d\mu(t)},
\end{equation*}
which satisfies $T\in[0,1]$ and $T=0$ if and only if $X\indpt Y\mid Z$, according to \citet[Theorem~2.1]{azadkia2019simple}.

Now consider an i.i.d. sample $(X_1,Y_1,Z_1), \dots, (X_n,Y_n,Z_n)$ from $p_*$. For each $i=1\dots,n$, let $N(i)$ be the index $j$ such that $Z_j$ is the nearest neighbor of $Z_i$ with respect to the Euclidean metric on $\bbR^{d_Z}$. Let $M(i)$ be the index $j$ such that $(X_j,Z_j)$ is the nearest neighbor of $(X_i,Z_i)$ in $\bbR^{d_X+d_Z}$. Let $R_i$ be the rank of $Y_i$. The RCI score is 
\begin{equation*}
	\hat{T}_n(X,Y|Z)=\frac{\sum_{i=1}^n(\min(R_i,R_{M(i)})-\min(R_i,R_{N(i)}))}{\sum_{i=1}^n(R_i-\min(R_i,R_{N(i)}))},
\end{equation*}
which is a consistent estimator of $T(X,Y|Z)$, according to \citet[Theorem~2.2]{azadkia2019simple}.

\section{Experimental Details}\label{app:detail}

\subsection{Implementations of baseline methods}

All baseline methods were run with the publicly available code from the authors' websites as listed below, expect KGV which we implemented by ourselves:
\begin{itemize}
\item GES: We adopt the FGES \citep{ramsey2017million} implementation from \url{https://github.com/eberharf/fges-py}. Note that all the methods using GES as the search procedure, including our proposed NCD, the adopted RCD, as well as the previous BIC and KGV, are based on the same implementation for searching with the only difference being the updating rule at each step. NCD and RCD follow the reframed GES update step in Algorithms \ref{FESupdate} and \ref{BESupdate}; BIC and KGV follow the standard GES update.
\item BIC: The linear-Gaussian BIC score is included in the above FGES implementation.
\item KGV: It adopted a Gaussian kernel with kernel width equal to twice of median distance between points in input space.
\item PC: An implementation is available through the {\tt py-causal} package at \url{https://github.com/bd2kccd/py-causal}. We choose SEM-BIC test with significance level $0.05$ for PC.
\item GSF: An implementation is available at the first author's github repository \url{https://github.com/Biwei-Huang/Generalized-Score-Functions-for-Causal-Discovery}.
\item CAM: An implementation is available through the CRAN R package repository at \url{https://cran.r-project.org/web/packages/CAM}.
\item NOTEARS: The code is available at the first author's github repository \url{https://github.com/xunzheng/notears}. 
\item DAG-GNN: The code is available at the first author's github repository \url{https://github.com/fishmoon1234/DAG-GNN}.
\item GraN-DAG: The code is available at the first author's github repository \url{https://github.com/kurowasan/GraN-DAG}. 
\end{itemize}

In the experiments, we mostly used the default hyperparameters found in the authors' codes unless otherwise stated.



\subsection{Experimental details and hyperparameters}

Since our model is based on deep neural networks (NNs), it is sensitive to the choice of hyperparameters, which is also observed in other neural network based causal discovery methods such as \citet{lachapelle2019gradient}. The hyperparameters in our NCD method include the threshold $\tau$ to control the sparsity level (number of edges) of the learned structure, the learning rates of the optimization steps in Algorithm \ref{alg}, and the neural network architectures (i.e., the number of hidden layers and hidden neurons per layer) for the test functions and nonlinear regressors. 
The principle of tuning $\tau$ is that a larger $\tau$ leads to a sparser DAG. To tune $\tau$, one needs an initial guess of the true sparsity, e.g., from domain expert knowledge, and tunes down $\tau$ if the learned DAG is much sparser than expected and vice versa. 

We use multilayer perceptrons (MLPs) to represent the test functions and regressors. A test function MLP has several blocks each of which consists of a fully connected layer and a ReLU activation function; a regressor MLP further adds batch normalization before the ReLU layer in each block. We adopt spectral normalization \citep{miyato2018spectral} in all networks to guarantee the Lipschitz continuity of them. The neural network models and optimization are implemented based on Pytorch. 
We use Adam optimizer with full batch gradients and a learning rate of 0.01 for both test functions and regressors. We take the training steps $T_t=20$ and $T_r=5$ for test functions and regressors respectively. 
Since test functions serve as transformations to detect correlation (which is a simpler task) while regressors need to fit the data (which is a more complex task), we keep the architecture of test functions fixed with 2 layers and 20 neurons per layer, while only tune the network size of regressors for different data. 
The above listed hyperparameters turn out to be very robust across different settings so we keep them unchanged across all settings. For different ground-truth causal models with varying dimensions and degrees, we only tune the threshold $\tau$ and the network depth and width. 

Roughly speaking, as we have more nodes and edges, we need larger NNs with more layers and neurons per layer. We suggest that practitioners tune the architecture on synthetic data with the same number of nodes and edges (roughly) and transfer the hyperparameters to the datasets at hand. The global score proposed below in \eqref{eq:global_score} can serve as a metric to evaluate each set of hyper-parameter values: a good set of hyper-parameter values should be the one that yields a low score (approaching 0) for the true structure and higher scores for any fake structures. In our experiments, we tune the hyperparameters on synthetic data sampled from additive noise models and transfer them to the PNL datasets, etc. Moreover, when some prior knowledge on the ground-truth structure is available, such as the absence or presence of a few edges and their orientations (which may imply some conditional independence conditions), we suggest tuning the hyper-parameters to match the prior information as much as possible. 

We also proposed in the paper an implementation of the reframed GES with the RCD measure in the literature, which involves no NN hyper-parameters and performs reasonably well across various settings. The only hyperparameter of the RCD implementation is the threshold $\tau$. In other words, when using the reframed GES in practice, there is also a good option that does not involve much hyper-parameter tuning, with some loss of accuracy in certain settings in comparison to the NN implementation NCD. 
We listed the specific hyperparameters for the our experiments in Table \ref{tab:hyperp}.

\begin{table}[t]
\centering
\begin{tabular}{ccccc}
\toprule
Setting & {depth} & {width} & {$\tau_{\text{NCD}}$} & {$\tau_{\text{RCD}}$} \\\midrule
PNL data with degree 2 & 3 & 40 & 0.005 & 0.05 \\
PNL data with degree 8 & 3 & 80 & 0.0001 & 0.001 \\
Multi-dimensional data & 3 & 50 & 0.01 & - \\
SynTReN & 4 & 100 & 0.3 & 0.5 \\ \bottomrule
\end{tabular}
\caption{Hyperparameters of NCD and RND for all settings.}\label{tab:hyperp}
\end{table}


Our NCD computation involves randomness coming from the neural network initialization (and stochastic optimization if adopted). 
Next, we introduce a metric based on the proposed NCD estimator to select among the random runs and to some extent guide hyperparameter tuning in an unsupervised manner (i.e., without access to the ground-truth structure). Given a candidate DAG $\cG$, let $\pa^\cG_i$ and $\nd^\cG_i$ be the sets of parents and non-descendants of node $X_i$, respectively. 
We propose the following global score to characterize how well the observational data satisfies the conditional independence relations entailed by $\cG$: 
\begin{equation}\label{eq:global_score}
    S_g(\cG)=\frac{1}{d}\sum_{i=1}^d \hat{S}_n(X_i,\nd^\cG_i|\pa^\cG_i).
\end{equation}

Apparently we have $S_g(\cG)\in[0,1]$. 
According to Theorems \ref{thm:pop_equiv} and \ref{thm:nci_cons}, we have $S_g(\cG)\pto0$ as $n\to\infty$ if and only if $\cG$ satisfies the Markov condition to the data distribution $P_\bV$. Hence a candidate DAG $\cG$ with a smaller global score $S_g(\cG)$ is regarded as a better estimate in the large sample limit. For each data set in our experiments, we run our reframed GES algorithm with NCD with two different random initializations and select the one with the lower global score.



\section{Additional Experimental Results}\label{app:add_exp}

We present the results of SID on the PNL data sets in Tables~\ref{tab:pnl2_sid}-\ref{tab:pnl8_sid} as a supplement to Tables~\ref{tab:pnl2_shd_f1}-\ref{tab:pnl8_shd_f1}. We can see that the SIDs are mostly consistent with the SHDs and F1 scores. In general, our NCD or RCD is among the best SID methods. In the sparse PNL-MULT data where GSF is the best with a smaller sample; in the dense graph (with degree 8), CAM performs well on both GP and MULT with a larger sample. 

Moreover, we present the results of all methods on the PNL data sets with 20 nodes, 2 expected degrees and 5000 samples in Tables \ref{tab:pnl_gp_d20}-\ref{tab:pnl_mult_d20}. We observe that our reframed GES with NCD or RCD performs among the best methods in this setting.

\begin{table*}
    \centering
    \begin{tabular}{lllll}
    \toprule
        Method & \multicolumn{1}{c}{GP (1k)} & \multicolumn{1}{c}{GP (5k)} & \multicolumn{1}{c}{MULT (1k)} & \multicolumn{1}{c}{MULT (5k)}  \\
        \midrule
        NCD   & \textbf{[11.2$\pm$7.0, 24.6$\pm$11.4]} & \textbf{[6.8$\pm$4.8, 19.2$\pm$9.5]} & [10.6$\pm$5.7, 23.6$\pm$10.6] & {[12.8$\pm$5.7, 25.4$\pm$6.1]} \\
        RCD   & [18.6$\pm$4.7, 30.2$\pm$7.5] & [17.4$\pm$2.9, 26.6$\pm$6.5] & [14.0$\pm$5.6, 29.8$\pm$11.3] & \textbf{[4.8$\pm$1.4, 22.0$\pm$7.8]} \\
        PC    & [17.0$\pm$6.6, 27.2$\pm$6.3] & [15.6$\pm$7.4, 27.2$\pm$7.9] & [18.4$\pm$8.3, 32.6$\pm$6.9] & [8.0$\pm$5.7, 23.2$\pm$5.9] \\
        BIC   & [15.0$\pm$8.8, 24.8$\pm$7.9] & [15.2$\pm$8.6, 23.4$\pm$8.4] & [7.6$\pm$7.3, 23.4$\pm$5.3] & [10.0$\pm$8.1, 25.2$\pm$4.7] \\
        KGV   & [18.5$\pm$4.0, 27.5$\pm$6.2] & [15.5$\pm$4.4, 29.0$\pm$2.2] & [15.3$\pm$6.1, 30.3$\pm$11.1] & [10.0$\pm$3.0, 28.6$\pm$10.4] \\
        CAM   & [10.8$\pm$6.2, 22.6$\pm$9.6] & [17.0$\pm$8.9, 28.6$\pm$9.8] & [27.6$\pm$13.6, 43.0$\pm$13.9] & [22.0$\pm$12.3, 37.2$\pm$16.5] \\
        NOTEARS & [21.2$\pm$3.4, 26.2$\pm$5.0] & [21.0$\pm$3.6, 26.6$\pm$4.7] & [15.0$\pm$4.2, 21.4$\pm$15.9] & [12.8$\pm$5.9, 17.0$\pm$9.1] \\
        DAG-GNN & [23.6$\pm$6.9, 27.4$\pm$5.8] & [30.4$\pm$11.0, 36.0$\pm$9.2] & [16.0$\pm$5.6, 23.2$\pm$9.9] & [16.0$\pm$3.9, 30.0$\pm$11.8] \\
        GraN-DAG & [27.0$\pm$7.5, 38.2$\pm$8.4] & [31.4$\pm$8.5, 41.8$\pm$7.3] & [14.0$\pm$5.7, 27.0$\pm$7.0] & [12.4$\pm$8.6, 25.6$\pm$6.5] \\
        GSF   & [14.2$\pm$8.6, 24.4$\pm$8.9] & - & \textbf{[5.0$\pm$1.4, 21.6$\pm$6.7]} & - \\
        \bottomrule
    \end{tabular}
    \caption{SID on PNL datasets with 10 nodes, 2 expected degrees, and 1000 and 5000 samples.}\label{tab:pnl2_sid}
\end{table*}

\begin{table*}
    \centering
    \begin{tabular}{lllll}
    \toprule
        Method & \multicolumn{1}{c}{GP (1k)} & \multicolumn{1}{c}{GP (5k)} & \multicolumn{1}{c}{MULT (1k)} & \multicolumn{1}{c}{MULT (5k)}  \\
        \midrule
        NCD   & \textbf{[56.6$\pm$11.5, 67.6$\pm$3.9]} & \textbf{[58.8$\pm$8.2, 66.0$\pm$3.7]} & \textbf{[59.2$\pm$10.0, 68.8$\pm$3.7]} & \textbf{[51.4$\pm$7.6, 69.0$\pm$3.9]} \\
        RCD   & [75.4$\pm$5.8, 75.4$\pm$5.8] & [73.6$\pm$4.3, 74.4$\pm$3.8] & [67.8$\pm$14.1, 77.0$\pm$4.5] & [53.2$\pm$6.6, 73.2$\pm$4.8] \\
        PC    & [78.2$\pm$10.8, 85.0$\pm$4.6] & [76.6$\pm$7.7, 82.4$\pm$3.6] & [72.2$\pm$4.5, 80.8$\pm$5.8] & [69.4$\pm$11.5, 78.4$\pm$5.6] \\
        BIC   & [69.8$\pm$7.1, 73.2$\pm$8.9] & [68.0$\pm$7.9, 68.8$\pm$8.3] & [67.8$\pm$6.7, 78.2$\pm$3.1] & [69.8$\pm$9.5, 77.4$\pm$4.7] \\
        KGV   & [74.6$\pm$7.7, 83.0$\pm$5.3] & [77.0$\pm$4.4, 79.6$\pm$5.0] & [67.2$\pm$3.4, 89.6$\pm$0.8] & [66.4$\pm$9.4, 87.8$\pm$3.0] \\
        CAM   & [65.6$\pm$10.3, 78.6$\pm$4.0] & \textbf{[54.6$\pm$13.4, 75.6$\pm$7.8]} & [56.8$\pm$4.1, 83.2$\pm$3.5] & \textbf{[51.8$\pm$27.0, 83.0$\pm$3.2]} \\
        NOTEARS & [75.8$\pm$1.8, 78.4$\pm$1.8] & [75.4$\pm$1.3, 78.0$\pm$1.2] & [63.4$\pm$6.9, 83.6$\pm$3.5] & [62.0$\pm$7.8, 83.6$\pm$4.3] \\
        DAG-GNN & [84.8$\pm$4.9, 89.6$\pm$0.5] & [86.8$\pm$2.6, 89.0$\pm$1.7] & [63.4$\pm$12.7, 83.2$\pm$3.0] & [72.4$\pm$5.4, 80.0$\pm$2.3] \\
        GraN-DAG & [72.6$\pm$22.8, 84.6$\pm$2.4] & [68.8$\pm$15.8, 78.6$\pm$5.6] & [67.4$\pm$6.6, 85.6$\pm$2.9] & [62.4$\pm$6.3, 82.6$\pm$3.7] \\
        GSF   & [69.6$\pm$10.3, 77.4$\pm$9.0] & - & [66.6$\pm$7.8, 81.0$\pm$3.8] & - \\
        \bottomrule
    \end{tabular}
    \caption{SID on PNL datasets with 10 nodes, 8 expected degrees, and 1000 and 5000 samples.}\label{tab:pnl8_sid}
\end{table*}

\begin{table*}
\centering
\begin{tabular}{lllll}
\toprule
Setting & \multicolumn{1}{c}{SHD} & \multicolumn{1}{c}{SID} & \multicolumn{1}{c}{F1}  \\
\midrule
NCD   & \textbf{9.2$\pm$4.1} & \textbf{[37.0$\pm$29.0,65.8$\pm$52.1]} & \textbf{0.69$\pm$0.10} \\
RCD   & 15.0$\pm$1.6 & [53.5$\pm$28.9, 89.2$\pm$34.7] & 0.45$\pm$0.09 \\
PC    & 13.2$\pm$1.9 & [42.6$\pm$15.0, 123.0$\pm$47.8] & 0.58$\pm$0.07 \\
BIC   & 11.6$\pm$1.8 & [53.8$\pm$16.1,93.6$\pm$22.9] & 0.60$\pm$0.03 \\
CAM   & \textbf{7.4$\pm$3.8} & \textbf{[38.8$\pm$21.5, 38.8$\pm$21.5]} & \textbf{0.75$\pm$0.11} \\
NOTEARS & 23.6$\pm$2.9 & [120.4$\pm$30.7, 120.4$\pm$30.7] & 0.06$\pm$0.02 \\
DAG-GNN & 21.4$\pm$3.4 & [105.2$\pm$35.5, 105.2$\pm$35.5] & 0.11$\pm$0.00 \\
GraN-DAG & 13.4$\pm$3.1 & [79.8$\pm$31.8, 79.8$\pm$31.8] & 0.50$\pm$0.14 \\
\bottomrule
\end{tabular}
\caption{Results on PNL-GP data with 20 nodes, 2 expected degrees and 5000 samples.}\label{tab:pnl_gp_d20}
\end{table*}

\begin{table*}
\centering
\begin{tabular}{lllll}
\toprule
Setting & \multicolumn{1}{c}{SHD} & \multicolumn{1}{c}{SID} & \multicolumn{1}{c}{F1}  \\
\midrule
NCD   & \textbf{10.2$\pm$1.9} & [30.4$\pm$4.2, 72.6$\pm$14.9] & 0.59$\pm$0.05 \\
RCD   & \textbf{8.2$\pm$1.6} & \textbf{[22.6$\pm$5.1, 69.2$\pm$12.3]} & \textbf{0.63$\pm$0.04} \\
PC    & 10.4$\pm$1.2 & [16.6$\pm$6.7,88.4$\pm$18.5] & 0.59$\pm$0.03 \\
BIC   & 12.0$\pm$2.3 & [24.8$\pm$6.2, 65.8$\pm$13.4] & 0.58$\pm$0.05 \\
CAM   & 22.0$\pm$3.5 & [99.8$\pm$12.5, 99.8$\pm$12.5] & 0.20$\pm$0.08 \\
NOTEARS & 29.0$\pm$2.3 & [52.8$\pm$13.4, 52.8$\pm$13.4] & 0.41$\pm$0.09 \\
DAG-GNN & 36.4$\pm$13.1 & [58.2$\pm$14.6, 58.2$\pm$14.6] & 0.35$\pm$0.09 \\
GraN-DAG & 18.2$\pm$3.86 & [75.4$\pm$10.1, 75.4$\pm$10.1] & 0.25$\pm$0.10 \\
\bottomrule
\end{tabular}
\caption{Results on PNL-MULT data with 20 nodes, 2 expected degrees and 5000 samples.}\label{tab:pnl_mult_d20}
\end{table*}

In addition to the performance in causal discovery, we also compare the computational time of different methods. We consider four methods related to the GES algorithm: the standard GES with the BIC score (BIC), the standard GES with GSF score (GSF), our reframed GES with the RCD and NCD score. Table~\ref{tab:time} reports the average running time of each method. 
As mentioned in the main text, our proposed NCD measure has an advantage over the kernel-based GSF in computation, both of which are nonparametric causal discovery methods. It has been well acknowledged that kernel methods suffer from high sample complexity (although some more efficient approximations exist), while neural networks can benefit from a large sample size without a severe compromise in computational time. We see from Table~\ref{tab:time} that NCD is significantly more computationally efficient than GSF, especially on datasets with a sparse graph structure. Moreover, since the NCD estimator is obtained by applying Algorithm~\ref{alg}, it is much more computationally demanding than score functions with an explicit formula to be easily computed such as BIC and RCD. This is clearly verified by the results in Table~\ref{tab:time}. RCD, as a nonparametric measure, costs more to compute than the simple BIC score, but is still fairly fast compared with the other two.

In fact, it is a trade-off between computational complexity and the quality of statistical estimation. As we noted in the main text, BIC is consistent only in restrictive parametric cases; otherwise, there exists a systematic error due to model misspecification, leading usually to poor results. This is verified by extensive results in causal discovery shown above and in the main text. In contrast, our NCD estimator is consistent in nonparametric settings which is much more general and flexible. Therefore, in applications of causal discovery where the accuracy of the estimation matters more than the computational cost, our approach has advantages over BIC.


\begin{table}[]
    \centering
    \begin{tabular}{lllllll}
    \toprule
    Dataset & Node & Degree & BIC & RCD & NCD & GSF  \\
    \midrule
    PNL-GP & 10 & 2 & $<1$ & $<1$ & 54.8 & 1195.2 \\
    PNL-MULT & 10 & 2 & $<1$ & 1.2 & 305 & 2122.8  \\
    PNL-GP & 10 & 8 & $<1$ & 3.8 & 748.8 & 1801.2 \\
    PNL-MULT & 10 & 8 & $<1$ & 9.2 & 618.0 & 1854.0 \\
    PNL-GP & 20 & 2 & $<1$ & 2.0 & 253.2 & 6461.4 \\
    PNL-MULT & 20 & 2 & $<1$ & 3.66 & 194.4 & 6380.4 \\
    \bottomrule
    \end{tabular}
    \caption{Average running time (seconds).}
    \label{tab:time}
\end{table}


\end{document}